\def\year{2020}\relax
\documentclass[letterpaper]{article} 
\usepackage{aaai20}  
\usepackage{times}  
\usepackage{helvet} 
\usepackage{courier}  
\usepackage[hyphens]{url}  
\usepackage{graphicx} 
\usepackage{graphicx}  
\usepackage{amsmath,amssymb,amsfonts}
\usepackage{booktabs,tabularx}

\title{Integrating Graph Contextualized Knowledge into Pre-trained Language Models}
\author{Bin He\textsuperscript{\rm 1}, Di Zhou\textsuperscript{\rm 1}, Jinghui Xiao\textsuperscript{\rm 1}, Xin Jiang\textsuperscript{\rm 1}, Qun Liu\textsuperscript{\rm 1}, Nicholas Jing Yuan\textsuperscript{\rm 2}, Tong Xu\textsuperscript{\rm 3}\\   
\textsuperscript{\rm 1}Huawei Noah’s Ark Lab\\
\textsuperscript{\rm 2}Huawei Cloud \& AI \\
\textsuperscript{\rm 3}School of Computer Science, University of Science and Technology of China\\ 
\{hebin.nlp, zhoudi7, xiaojinghui4, jiang.xin, qun.liu, nicholas.yuan\}@huawei.com, tongxu@ustc.edu.cn 
}

\begin{document}

\maketitle

\begin{abstract}
%
Complex node interactions are common in knowledge graphs, and these interactions also contain rich knowledge information.
However, traditional methods usually treat a triple as a training unit during the knowledge representation learning (KRL) procedure, neglecting contextualized information of the nodes in knowledge graphs (KGs).
We generalize the modeling object to a very general form, which theoretically supports any subgraph extracted from the knowledge graph, and these subgraphs are fed into a novel transformer-based model to learn the knowledge embeddings.
To broaden usage scenarios of knowledge, pre-trained language models are utilized to build a model that incorporates the learned knowledge representations.
Experimental results demonstrate that our model achieves the state-of-the-art performance on several medical NLP tasks, and improvement above TransE indicates that our KRL method captures the graph contextualized information effectively.

\end{abstract}

\section{Introduction}

Pre-trained language models learn contextualized word representations on large-scale text corpus through an self-supervised learning method, which are finetuned on downstream tasks and can obtain the state-of-the-art (SOTA) performance \cite{peters2018deep,radford2018improving,devlin2019bert}.
This gradually becomes a new paradigm for natural language processing research.
Recently, knowledge information has been integrated into pre-trained language models to enhance the language representations, such as ERNIE-Tsinghua \cite{zhang-etal-2019-ernie} and ERNIE-Baidu \cite{sun2019ernie}.
\citeauthor{zhang-etal-2019-ernie} \shortcite{zhang-etal-2019-ernie} makes a preliminary attempt to utilize knowledge information learned from a knowledge graph to improve the performance of some knowledge-driven tasks. 
Intuitively, this model can be directly applied to the medical domain, where large-scale corpora and knowledge graphs are both available.

\begin{figure}[!t]
	\centering
	\includegraphics[width=\linewidth]{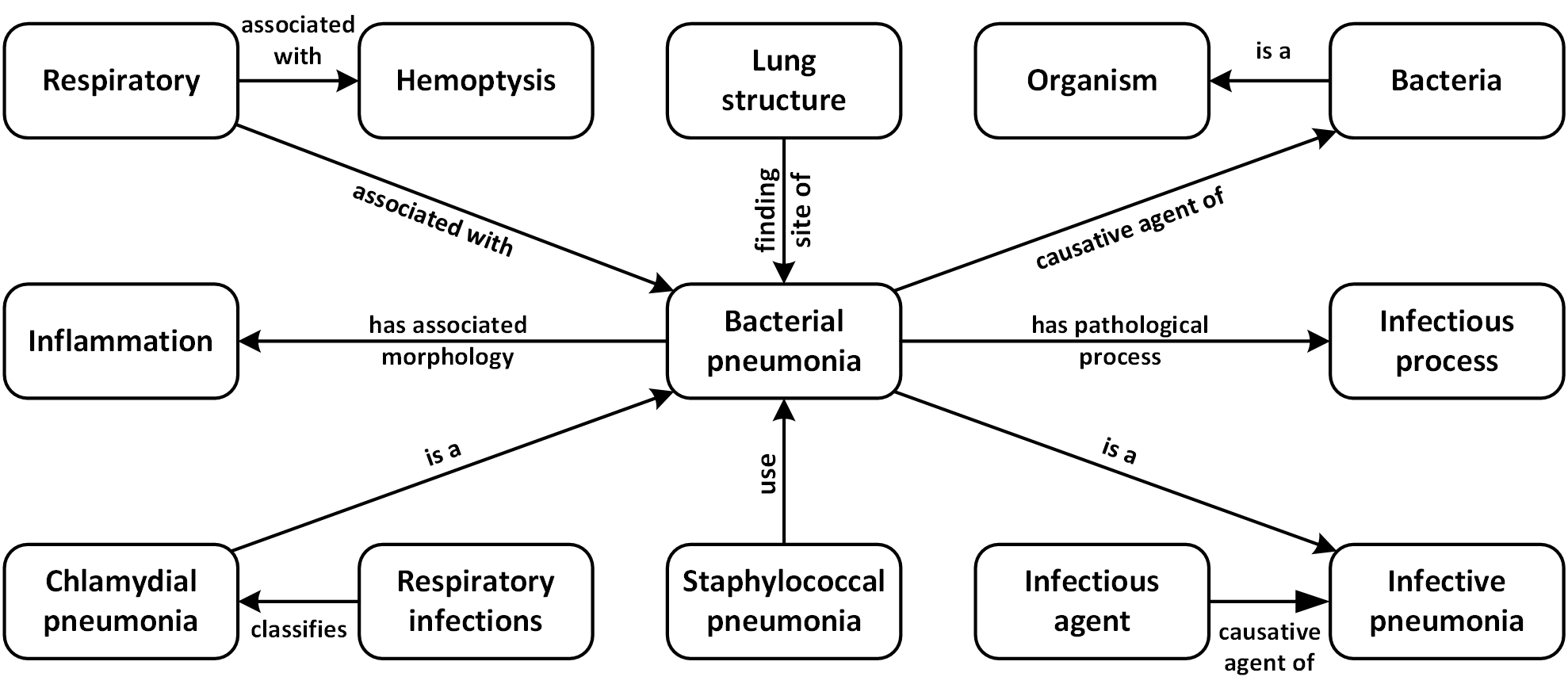}
	\caption{\label{fig:kg} A subgraph extracted from a medical knowledge graph. The rectangles represent medical entities and directed arrows denote the relations between these entities. Only part of the neighbor nodes are listed for clarity.}
\end{figure}

In ERNIE-Tsinghua, entity embeddings are learned by TransE \cite{bordes2013translating}, which is a popular transition-based method for knowledge representation learning (KRL).
TransE simply considers each triple in the knowledge graph as a training instance, which may be insufficient to model complex information transmission between nodes in the knowledge graph.
In the medical knowledge graph, some entities have a large number of related neighbors, and TransE can not model all the neighbors for the corresponding entity simultaneously.
Figure~\ref{fig:kg} shows a subgraph of a medical knowledge graph containing several medical entities.
In this figure, four incoming and four outgoing neighboring nodes (hereinafter called ``in-entity" and ``out-entity") of node ``\textit{Bacterial pneumonia}" are listed with various types of relation between the nodes.
Therefore, in order to learn more comprehensive node embeddings, it is necessary to incorporate more interactive information between nodes (we call this ``graph contextualized information'').

Graph attention networks (GATs) \cite{velickovic2018graph} update the entity embedding by its 1-hop neighbors, which pays more attention to the node information interactions.
Based on GATs, multi-hop neighbors of a given entity are integrated to capture further information \cite{nathani-etal-2019-learning}.
Inspired by previous work, we extend the information interaction between nodes to a more general approach, which can treat arbitrary subgraph from the knowledge graph as a training example.
Specifically, the subgraph is transformed into a sequence of nodes, and the KRL procedure is performed in a way similar to training a language model. In this manner, more comprehensive contextualized information of each node can be incorporated into the learned knowledge representations.
Besides, we believe that entities and relations should be mutually influential  during the KRL process, so the relations are regarded as graph nodes as well and learned jointly with entities.

Finally, our pre-trained model called BERT-MK (a \textbf{BERT}-based language model with \textbf{M}edical \textbf{K}nowledge) is learned with a large-scale medical corpus, and integrated with the medical knowledge graph represented through the aforementioned KRL approach.
Our contributions are as follows:
\begin{itemize}
	\item A novel knowledge representation learning method that is capable of modeling arbitrary subgraph is proposed. This method greatly enriches the amount of information in the knowledge representation and explores the joint learning of entities and relations.
	\item Graph contextualized knowledge is integrated to enhance the performance of pre-trained language models, which outperforms state-of-the-art models on several NLP tasks in the medical domain.
\end{itemize}

%

\section{Methodology}

In our model, each time a certain number of nodes and their linked nodes are selected from the knowledge graph to construct a training sample.
Then, the embeddings of these nodes are learned from their neighbors by a novel knowledge representation learning algorithm based on Transformer.
Finally, the learned knowledge representation is integrated into the language model to enhance the model pre-training and fine-tuning.

\subsection{Learning Knowledge Graph Embeddings by Transformer}
\label{sec:kg}

Transformer \cite{vaswani2017attention} can be used as a powerful encoder to model the sequential inputs.
Recently,  \citeauthor{koncel2019text} \shortcite{koncel2019text} extend Transformer to encode graph-structured inputs.
In their work, text is first transformed into a graph, which is then encoded by the graph Transformer and fed into a text generation model.
Inspired by their work, we convert a knowledge subgraph into a sequence of nodes, and utilize a Transformer-based model to learn the node embeddings. We call this model ``KG-Transformer", where ``KG'' denotes ``knowledge graph'').
More details of the method are described in the following subsections.

\subsubsection{Graph Conversion}
We denote a knowledge graph as $\mathcal{G}=(\mathcal{E}, \mathcal{R})$, where $\mathcal{E}$ represents the entity set and $\mathcal{R}$ is the set of relations between the pairs of entities in $\mathcal{G}$. 
A triple in $\mathcal{G}$ is denoted by $t=(e_s, r, e_o)$, where $e_s$ is a subjective entity, $e_o$ is an objective entity, and $r$ is the relation between $e_s$ and  $e_o$.
Figure~\ref{fig:kg} gives an example of a subgraph of the medical KG.
Two entities (rectangles) and a relation (arrow) between them constructs a knowledge triple, for example, (\textit{Bacterial pneumonia, causative agent of, Bacteria}).

\begin{figure}[!t]
	\centering
	\includegraphics[width=\linewidth]{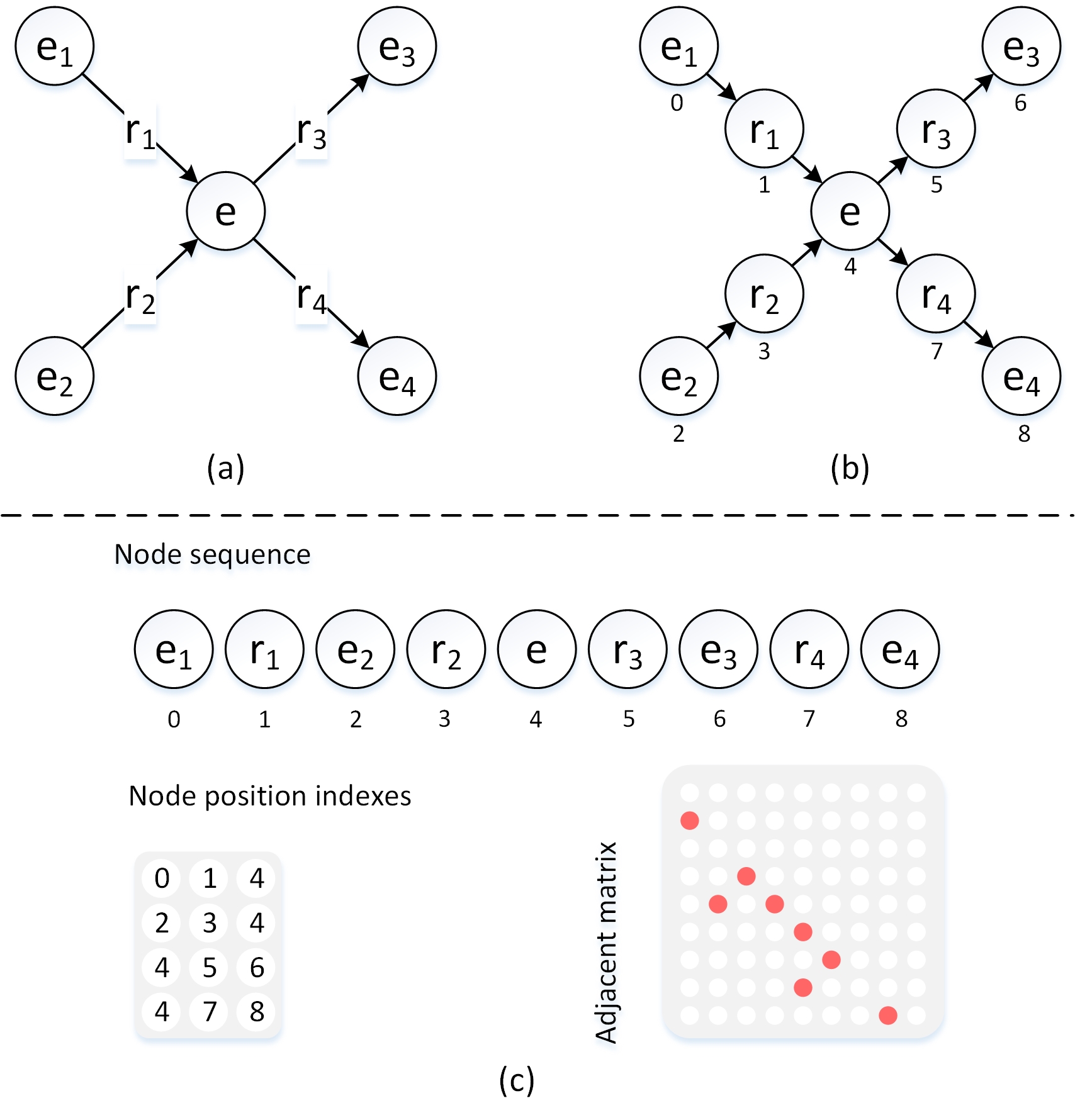}
	\caption{\label{fig:gn} Converting a subgraph extracted from the knowledge graph into the input of the KG-Transformer model. (a) $e$ refers to the entity, and $r$ represents the relation. (b) Relations are transformed into graph nodes, and all nodes are assigned a numeric index. (c) Each row in the matrix for node position indexes represents the index list for an triple in (b); the adjacent matrix indicates the connectivity (the red points equal to 1 and the white points are 0) between the nodes in (b).}
\end{figure}

\begin{figure*}[!t]
	\centering
	\includegraphics[width=\linewidth]{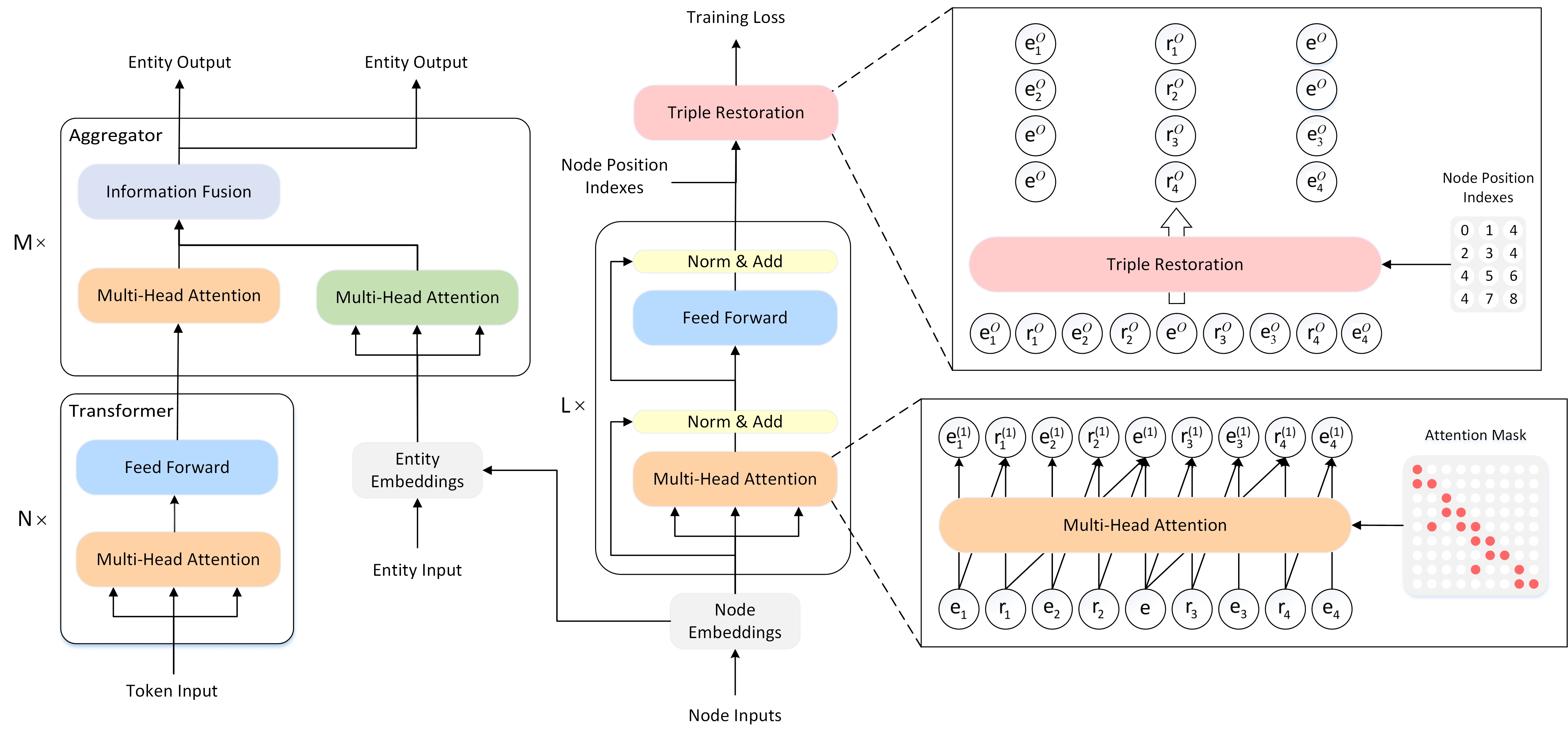}
	\caption{\label{fig:model} The model architecture of BERT-MK. The left part is the pre-trained language model, in which entity information learned from the knowledge graph is incorporated. The right part presents the KG-Transformer model, which utilizes the training sample illustrated in Figure~\ref{fig:gn} to describe the knowledge representation learning process. $e_1$, $e_1^{(1)}$, $e_1^O$ is the embedding of the input node, the updated node and the output node, respectively.}
\end{figure*}

Single triple is treated as a training sample in the traditional KRL methods, such as TransE \cite{bordes2013translating} and ConvKB \cite{nguyen2018novel}. 
In this setting, information from the entity's neighbors cannot be involved to update the entity embedding simultaneously. Graph attention networks (GATs) \cite{velickovic2018graph} are proposed to solve this problem.
\citeauthor{nathani-etal-2019-learning} \shortcite{nathani-etal-2019-learning} utilize GAT to build a KRL model, in which all the subjective entities directing to an objective entity are used to learn the embedding of the objective entity.
In our work, we propose a more general method that takes any subgraph of the KG as a training sample, which greatly enriches the contextualized information in learning the knowledge representation.
For ease of explanation,  in Figure~\ref{fig:gn}(a), we choose one entity with its two in-entities and out-entities to build a training sample.

Besides, relations in the KG are also learned as nodes equivalent to entities in our model, which achieves joint training of entity embeddings and relation embeddings.
The node conversion process is illustrated in Figure~\ref{fig:gn}(b). The knowledge graph can be redefined as $G=(V, E)$, where $V$ represents the nodes in $G$, involving entities in $\mathcal{E}$ and relations in $\mathcal{R}$, and $E$ denotes a adjacency matrix indicating the directed edges among the nodes in $V$.
The adjacency matrix in Figure~\ref{fig:gn}(c) shows the connectivity between the nodes in Figure~\ref{fig:gn}(b).

The conversion result of a subgraph is shown in Figure~\ref{fig:gn}(c), including a node sequence, a node position index matrix and an adjacency matrix.
Each row of the node position index matrix corresponds to a triple in the graph.
For example,  the triple $(e_1, r_1, e)$ is represented as the first row $(0,1,4)$ in this matrix.
In the adjacency matrix, the elements $a_{ij}$ equals 1 if the node $i$ is connected to node $j$ in Figure~\ref{fig:gn}(b), and 0 otherwise.

\subsubsection{Transformer-based Encoder}

We denote a node sequence as $\{x_1, \dots, x_N\}$, where $N$ is the length of the input sequence. Besides, a node position index matrix and an adjacency matrix are defined as $\mathbf{P}$ and $\mathbf{A}$, respectively. 
Entity embeddings and relation embeddings are integrated in a matrix $\mathbf{V}$, $\mathbf{V}\in\mathbb{R}^{(n_e+n_r)\times d}$, where $n_e$ is the entity number in $\mathcal{E}$ and $n_r$ is the relation type number in $\mathcal{R}$.
The node embeddings $\mathbf{X}=\{\mathbf{x}_1, \dots, \mathbf{x}_N\}$ can be generated by looking up node sequence $\{x_1, \dots, x_N\}$ in embedding matrix $\mathbf{V}$.
$\mathbf{X}$, $\mathbf{P}$ and $\mathbf{A}$ constitute the input of our KRL model, as shown in Figure~\ref{fig:model}.

The inputs are fed into a Transformer-based model to encode the node information.
\begin{equation}
\mathbf{x}_i' =\bigoplus_{h=1}^H\sum_{j=1}^{N}{\alpha_{ij}^h\cdot(\mathbf{x}_j\cdot \mathbf{W}_\mathrm{v}^h)},
\end{equation}
\begin{equation}
\alpha_{ij}^h=\frac{\exp(a_{ij}^h)}{\sqrt{d/H}\cdot\sum_{n=1}^{N}\exp(a_{in}^h)},
\end{equation}
\begin{equation}
\label{eq:mask}
a_{ij}^h=\mathtt{Masking}((\mathbf{x}_i\cdot \mathbf{W}_\mathrm{q}^h)\cdot(\mathbf{x}_j\cdot \mathbf{W}_\mathrm{k}^h)^\mathrm{T}), \mathbf{A}_{ij}+\mathbf{I}_{ij}),
\end{equation}
where $\mathbf{x}_i'$ is the new embedding for node $x_i$.
$\bigoplus$ denotes the concatenation of the $H$ attention heads in this layer, $\alpha_{ij}^h$ and $\mathbf{W}_\mathrm{v}^h$ are the attention weight of node $x_j$ and a linear transformation of node embedding $\mathbf{x}_j$ in the $h^\mathtt{th}$ attention head, respectively.
The $\mathtt{Masking}$ function in Equation \ref{eq:mask} restraints the contextualized dependency among the input nodes, only the \textit{degree-in} nodes and the current node itself are involved to update the node embedding.
Similar to $\mathbf{W}_\mathrm{v}^h$, $\mathbf{W}_\mathrm{q}^h$ and $\mathbf{W}_\mathrm{k}^h$ are independent linear transformations of node embeddings.
Then, the new node embeddings are fed into the feed forward layer for further encoding.

As in Transformer model, we stack the aforementioned Transformer blocks $L$ times. The output of the Transformer-based encoder can be formalized as 
\begin{equation}
\mathbf{X}^O=\{\mathbf{x}_1^O, \dots, \mathbf{x}_N^O\}.
\end{equation}

\subsubsection{Training Objective}

The output of the encoder $\mathbf{X}^O$ and the node position indexes $\mathbf{P}$ are utilized to restore the triples and generate the embeddings of these triples:

\begin{equation}
\mathbf{T}=\mathtt{TripleRestoration}(\mathbf{X}^O,\mathbf{P}),
\end{equation}
where $\mathbf{T}_k=(\mathbf{x}_{e^k_s}, \mathbf{x}_{r^k}, \mathbf{x}_{e^k_o})$ and $\mathbf{P}_k=(e^k_s, r^k, e^k_o)$ is the position index of a valid knowledge triple.

In this study, the translation-based scoring function \cite{han2018openke} is adopted to measure the energy of a knowledge triple.
The node embeddings are learned by minimizing a margin-based loss function on the training data:
\begin{equation}
\mathcal{L}=\sum_{\mathbf{t}\in\mathbf{T}}\mathtt{max}\{d(\mathbf{t})-d(f(\mathbf{t}))+\gamma, 0\},
\end{equation}
where $\mathbf{t}=(\mathbf{t}_s,\mathbf{t}_r,\mathbf{t}_o)$, $d(\mathbf{t})=|\mathbf{t}_s+\mathbf{t}_r-\mathbf{t}_o|$, $\gamma>0$ is a margin hyperparameter, $f(\mathbf{t})$ is an entity replacement operation that the head entity or the tail entity in a triple is replaced and the replaced triple is an invalid triple in the KG.

\subsection{Integrating Knowledge into the Language Model}
\label{sec:lm}

Given a comprehensive medical knowledge graph, graph contextualized knowledge representations can be learned using the KG-Transformer model.
We follow the language model architecture proposed in \cite{zhang-etal-2019-ernie}, graph contextualized knowledge is utilized to enhance the medical language representations.
The language model pre-training process is shown in the left part of Figure~\ref{fig:model}.
The Transformer block encodes word contextualized representation while the aggregator block implements the fusion of knowledge and language information.

According to the characteristics of medical NLP tasks, domain-specific finetuning procedure is designed.
Similar to BioBERT \cite{lee2019biobert}, symbol ``@'' and ``\$'' are used to mark the entity boundary, which indicate the entity positions in a sample and distinguish different relation samples sharing the same sentence.
For example, the input sequence for the relation classification task can be modified into ``\textit{[CLS] pain control was initiated with morphine but was then changed to @ \textit{demerol} \$, which gave the patient better relief of @ \textit{his epigastric pain} \$}".
In the entity typing task, entity mention and its context are critical to predict the entity type, so more localized features of the entity mention will benefit this prediction process.
In our experiments, the entity start tag ``@'' is selected to represent an entity typing sample.

\section{Experiments}

\begin{table*}[!t]
	\centering
	\caption{\label{tab:umls} Statistics of UMLS.}\smallskip
	\begin{tabular}{ccccccc}
		\hline  & \bf \# Entities & \bf \# Relations & \bf \# Triples & \bf Ave. in-degree & \bf Ave. out-degree & \bf Median degree   \\ \hline
		UMLS & 2,842,735 & 874 & 13,555,037 & 5.05 & 5.05 & 4   \\ \hline 
	\end{tabular}
\end{table*}

\begin{table*}[!t]
	\centering
	\caption{\label{tab:dataset} Statistics of the datasets. Most of these datasets do not follow a standard train-valid-test set partition, and we adopt some traditional data partition ways to do model training and evaluation.}\smallskip
	\begin{tabular}{llcccccc}
		\hline \bf Task & \bf Dataset & \bf \# Train & \bf \# Valid & \bf \# Test  \\ \hline
		Entity Typing & 2010 i2b2/VA \cite{uzuner20112010} & 16,519 & - & 31,161 \\
		& JNLPBA  \cite{kim2004introduction} & 51,301 & - & 8,653 \\
		& BC5CDR \cite{li2016biocreative} & 9,385 & 9,593 & 9,809 \\
		Relation Classification & 2010 i2b2/VA \cite{uzuner20112010} & 10,233 & - & 19,115 \\
		& GAD \cite{bravo2015extraction} & 5,339 & - & - \\
		& EU-ADR \cite{van2012eu} & 355 & - & -  \\
		\hline 
	\end{tabular}
\end{table*}

\subsection{Dataset}

\paragraph{Medical Knowledge Graph}
The Unified Medical Language System (UMLS) \cite{bodenreider2004unified} is a comprehensive knowledge base in the biomedical domain, which contains large-scale concept names and relations among them.
The metathesaurus in UMLS involves various terminology  systems and comprises about 14 million terms covering  25 different languages. 
In this study, a subset of this knowledge base is extracted to construct the medical knowledge graph for KRL. Non-English and long terms are filtered, and the final statistics is shown in Table~\ref{tab:umls}.


\paragraph{Corpus for Pre-training}
To ensure that sufficient medical knowledge can be integrated into the language model, PubMed abstracts and PubMed Central full-text papers are chosen as the pre-training corpus, which are open-access datasets for biomedical and life sciences journal literature. 
Since sentences in different paragraphs may not have good context coherence, paragraphs are selected as the document unit for next sentence prediction.
The Natural Language Toolkit (NLTK) is utilized to split the sentences within a paragraph, and sentences having less than 5 words are discarded.
As a result, a large corpus containing 9.9B tokens is achieved for language model pre-training.

In our model, medical terms appearing in the corpus need to be aligned to the entities in the UMLS metathesaurus before pre-training.
To make sure the coverage of identified entities in the metathesaurus, the forward maximum matching (FMM) algorithm is used to extract the term spans from the corpus aforementioned, and spans less than 5 characters are filtered.
Then, BERT vocabulary is used to tokenize the input text into word pieces, and the medical entity is aligned with the first subword of the identified term.

\paragraph{Downstream tasks}

In this study, entity typing and relation classification tasks in the medical domain are used to evaluate the models.

\textbf{Entity Typing} Given a sentence with an entity mention tagged, the task of entity typing is to identify the semantic type of this entity mention.
For example, the type ``\textit{medical problem}" is used to label the entity mention in the sentence ``\textit{he had a differential diagnosis of $\langle\mathtt{e}\rangle$ asystole $\langle/\mathtt{e}\rangle$}".
To the best of our knowledge, there are no publicly available entity typing datasets in the medical domain, therefore, three entity typing datasets are constructed from the corresponding medical named entity recognition datasets.
Entity mentions and entity types are annotated in these datasets, in this study, entity mentions are considered as input while entity types are the output labels.
Table~\ref{tab:dataset} shows the statistics of the datasets for the entity typing task.

\textbf{Relation Classification} 
Given two entities within one sentence, the task aim is to determine the relation type between the entities.
For example, 
in sentence ``\textit{pain control was initiated with morphine but was then changed to $\langle\mathtt{e_1}\rangle$ demerol $\langle/\mathtt{e_1}\rangle$, which gave the patient better relief of $\langle\mathtt{e_2}\rangle$ \textit{his epigastric pain} $\langle/\mathtt{e_2}\rangle$}", the relation type between two entities is \textit{TrIP} (Treatment Improves medical Problem).
In this study, three relation classification datasets are utilized to evaluate our models, and the statistics of these datasets are shown in Table~\ref{tab:dataset}.

\subsection{Implementation Details}

\paragraph{Knowledge Representation Learning}
To achieve a basic knowledge representation, UMLS triples are fed into the TransE model.
The OpenKE toolkit \cite{han2018openke} is adopted to train the entity and relation embeddings, and the embedding dimension is set to 100 while training epoch number is set to 10000.

Following the initialization method used in \cite{nguyen2018novel,nathani-etal-2019-learning}, the embeddings produced by TransE are utilized to initialize the representation learning by the KG-Transformer model.
Both the layer number and the hidden head number is set to 4.
Due to the median degree of nodes in UMLS is 4 (shown in Table\ref{tab:umls}), one node with two in-nodes and two out-nodes is sampled as a training instance.
The KG-Transformer model runs 1200 epochs on a single NVIDIA Tesla V100 (32GB) GPU to train the knowledge embeddings, with a batch size of 50000.

\paragraph{Pre-training}

First,  a medical ERNIE (MedERNIE) model is trained on UMLS triples and the PubMed corpus, inheriting the same model hyperparameters used in \cite{zhang-etal-2019-ernie}.
Besides, the entity embeddings learned by the KG-Transformer model are integrated into the language model to train the BERT-MK model.
In our work, we align the same amount of pre-training with BioBERT, which uses the same pre-training corpus as ours, and finetune the BERT-Base model on the PubMed corpus for one epoch.

\paragraph{Finetune}

Since there is no standard development set in some datasets, we divide the training set into a new training set and a development set by 4:1.
For datasets containing a standard test set, we preform each experiment five times under specific experimental settings with different random seeds, and the average result is used to improve the evaluation reliability.
Besides, 10-fold cross-validation  method is used to evaluate the model performance for the datasets without a standard test set.
According to the maximum sequence length of the sentences in each dataset, the input sequence length for 2010 i2b2/VA \cite{uzuner20112010}, JNLPBA \cite{kim2004introduction}, BC5CDR \cite{li2016biocreative}, GAD \cite{bravo2015extraction} and EU-ADR \cite{van2012eu} are set to 390, 280, 280, 130 and 220, respectively.
The initial learning rate is set to 2e-5.

\begin{table*}[!t]
	\centering
	\caption{\label{tab:results} Experimental results on the entity typing and relation classification tasks. Accuracy (Acc), Precision, Recall, and F1 scores are used to evaluate the model performance. The results given in previous work are underlined. E-SVM is short for Ensemble SVM \cite{bhasuran2018automatic}, which achieves SOTA performance in GAD. CNN-M stands for CNN using multi-pooling \cite{he2019classifying}, and this model is the SOTA model in 2010 i2b2/VA.}\smallskip
	\begin{tabular}{llccccccc}
		\hline \bf Task & \bf Dataset & \bf Metrics & \bf E-SVM & \bf CNN-M & \bf BERT-Base & \bf BioBERT & \bf SCIBERT & \bf BERT-MK   \\ 
		\hline
		Entity  & 2010 i2b2/VA & Acc & - & - & 96.76 & 97.43 & \bf 97.74 &  97.70   \\
		Typing & JNLPBA & Acc & - & - & 94.12 & 94.37 & \bf 94.60  & 94.55 \\
		& BC5CDR & Acc & - & - & 98.78 & 99.27 & 99.38 & \bf 99.54 \\ 
		\cline{2-9}
		& Average & Acc & - & - & 96.55 & 97.02 & 97.24 & \bf 97.26 \\ 
		\hline
		Relation  & 2010 i2b2/VA  & P & - & \underline{73.1} & 72.6 & 76.1 & 74.8 & \bf 77.6 \\
		Classification & & R & - & \underline{66.7} & 65.7 & 71.3 & 71.6 & \bf 72.0 \\
		& & F & - & \underline{69.7} & 69.2 & 73.6 & 73.1 & \bf 74.7 \\
		& GAD & P & \underline{79.21} & - & \underline{74.28} & \underline{76.43} & 77.47 & \bf 81.67 \\
		& & R & \underline{89.25} & - & \underline{85.11} & \underline{87.65} & 85.94 & \bf 92.79 \\
		& & F & \underline{83.93} & - & \underline{79.33} & \underline{81.66} & 81.45 & \bf 86.87 \\
		& EU-ADR & P & - & - & \underline{75.4}5 & \underline{81.05} & 78.42 & \bf 84.43 \\
		& & R & - & - & \bf \underline{96.55} & \underline{93.90} & 90.09 & 91.17 \\
		& & F & - & - & \underline{84.71} & \underline{87.00} & 85.51 & \bf 87.49 \\ 
		\cline{2-9}
		& Average & P & - & -  & 74.11 & 77.86 & 76.90 & \bf 81.23 \\
		& & R &-  & - & 82.45 & 84.23 & 82.54 & \bf 85.32 \\
		& & F & -  & - & 77.75 & 80.75 & 80.02 & \bf 83.02 \\
		\hline 
	\end{tabular}
\end{table*}

\paragraph{Baselines}

In addition to the state-of-the-art models on these datasets, we have also added the popular BERT-Base model and another two models pre-trained on biomedical literature for further comparison.

\textbf{BERT-Base} \cite{devlin2019bert} This is the original bidirectional pre-trained language model proposed by Google, which achieves state-of-the-art performance on a wide range of NLP tasks.

\textbf{BioBERT} \cite{lee2019biobert} This model follows the same model architecture as the BERT-Base model, but with the PubMed abstracts and PubMed Central full-text articles (about 18B tokens) used to do model finetuning upon BERT-Base.

\textbf{SCIBERT} \cite{beltagy2019scibert} A new wordpiece vocabulary is built based on a large scientific corpus (about 3.2B tokens). 
Then, a new BERT-based model is trained from scratch using this new scientific vocabulary and the scientific corpus.
Since a large portion of the scientific corpus is biomedical articles, this scientific vocabulary can also be regarded as a biomedical vocabulary, which can improve the performance of downstream tasks in the biomedical domain effectively.

\subsection{Results}

Table ~\ref{tab:results} presents the experimental results on the entity typing and relation classification tasks.
For entity typing tasks, all these pre-trained language models achieve high accuracy, indicating that the type of a medical entity is not as ambiguous as that in the general domain.
BERT-MK outperforms BERT-Base, BioBERT and SCIBERT by 0.71\%, 0.24\% and 0.02\% on the average accuracy, respectively.
Without using external knowledge in the pre-trained language model, SCIBERT achieves comparable results to BERT-MK, which proves that a domain-specific vocabulary is critical to the feature encoding of inputs.
Long tokens are relatively common in the medical domain, and these tokens will be split into short pieces when a domain-independent vocabulary is used, which will cause an overgeneralization of lexical features.
Therefore, a medical vocabulary generated by the PubMed corpus can be introduced into BERT-MK in the following work.


On the relation classification tasks, BERT-Base does not perform as well as other models, which indicates that pre-trained language models require a domain adaptation process when used in restricted domains.
Compared with BioBERT, which utilizes the same domain-specific corpus as ours for domain adaptation of pre-trained language models, BERT-MK improves the average F score by 2.27\%, which demonstrates medical knowledge has indeed played a positive role in the identification of medical relations.
The following example provides a brief explanation of why medical knowledge improves the model performance of the relation classification tasks.
``\textit{On postoperative day number three , patient went into $\langle\mathtt{e_1}\rangle$ atrial fibrillation $\langle/\mathtt{e_1}\rangle$ , which was treated appropriately with $\langle\mathtt{e_2}\rangle$ metoprolol $\langle/\mathtt{e_2}\rangle$ and digoxin and converted back to sinus rhythm}" is a relation sample from the 2010 i2b2/VA dataset, and the relation label is \textit{TrIP}.
Meanwhile, the above entity pair can be aligned to a knowledge triple (\textit{atrial fibrillation}, \textit{may be treated by}, \textit{metoprolol}) in the medical knowledge graph.
Obviously, this knowledge information is advantageous for the relation classification of the aforementioned example.

\paragraph{TransE vs. KG-Transformer}

In order to give a more intuitive analysis of our new KRL method for the promotion of pre-trained language models, we compare MedERNIE (TransE is used to learn knowledge representation) and BERT-MK (corresponding to KG-Transformer) on two relation classification datasets.
Table~\ref{tab:ernie} demonstrates the results of these two models.
As we can see, integrating knowledge information learned by the KG-Transformer model, the performance increases the F score by 0.9\% and 0.64\% on two relation classification datasets, respectively, which indicates that improvement of the knowledge quality has a beneficial effect on the pre-trained language model.

\begin{figure}[!h]
	\centering
	\includegraphics[width=\linewidth]{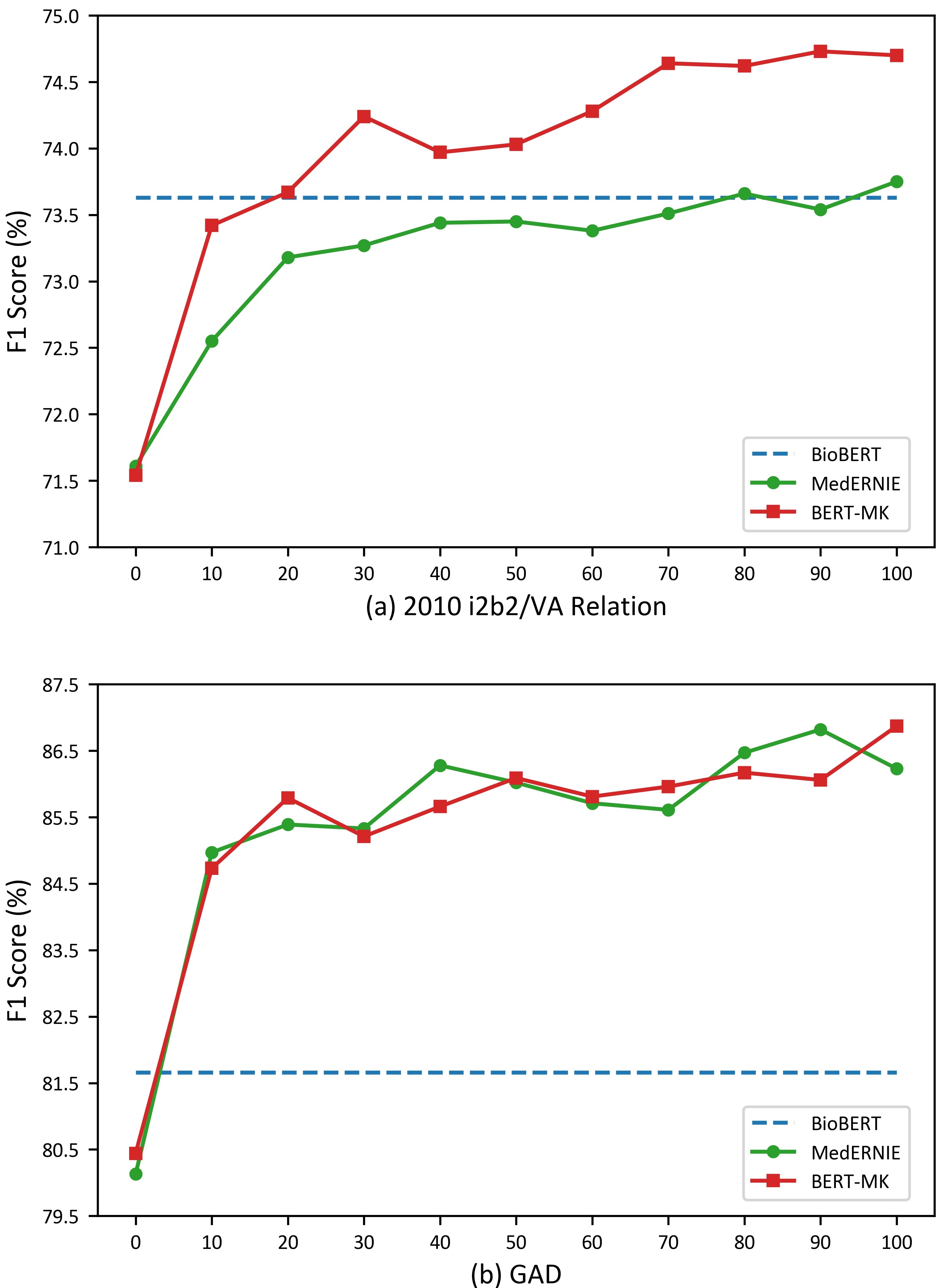}
	\caption{\label{fig:traindata} Model performance comparison with increasing amount of the pre-trained data. The x-axis represents the proportion of the medical data used for pre-training. 0 means no medical data is utilized, so the BERT-Base is used as an initialization parameter for the model finetuning. 100 indicates the model is pre-trained on the medical corpus for one epoch. BioBERT pre-trains on the PubMed corpus for one epoch, which is drawn with dashed lines in the figure as a baseline for comparison.}
\end{figure}

In Figure~\ref{fig:traindata}, as the amount of pre-training data increases, BERT-MK always outperforms MedERNIE on the 2010 i2b2/VA relation dataset, and the performance gap has an increasing trend. However, on the GAD dataset, the performance of BERT-MK and MedERNIE are intertwined.
We link the entities in the relation samples to the knowledge graph, and perform statistical analysis on the relationships between the linked nodes.
We observe that there are 136 2-hop neighbor relationships in 2010 i2b2/VA, while only 1 in GAD.
The second case shown in Table~\ref{tab:case} gives an example of the situation described above.
Triple (\textit{CAD, member of, Other ischemic heart disease (SMQ)}) and (\textit{Other ischemic heart disease (SMQ), has member, Angina symptom}) are found in the medical knowledge graph, which indicates entity \textit{cad} and entity \textit{angina symptoms} have a 2-hop neighbor relationship.
KG-Transformer learns these 2-hop neighbor relationships in 2010 i2b2/VA and produces an improvement for BERT-MK.
However, due to the characteristics of the GAD dataset, the capability of KG-Transformer is limited.

\begin{table}[!ht]
	\centering
	\caption{\label{tab:ernie} TransE vs. KG-Transformer. Since the EU-ADR dataset is too small, the model comparison on this dataset is not listed in this table.}\smallskip
	\begin{tabularx}{\linewidth}{lXXXXXX}
		\hline \bf Dataset & \multicolumn{3}{c}{\bf MedERNIE} & \multicolumn{3}{c}{\bf BERT-MK}   \\
		\cmidrule(r){2-4} \cmidrule(l){5-7}
		& P & R & F & P & R & F \\ 
		\hline
		2010 i2b2/VA & 76.6 & 71.1 & 73.8 & 77.6 & 72.0 & 74.7 \\
		GAD & 81.28 & 91.86 & 86.23 & 81.67 & 92.79 & 86.87 \\
		\hline
		Average & 78.94 & 81.48 & 80.02 & \bf 79.64 & \bf 82.40 & \bf 80.79 \\
		\hline 
	\end{tabularx}
\end{table}

\begin{table*}[!t]
	\centering
	\scriptsize
	\caption{\label{tab:case} Case study on the 2010 i2b2/VA relation dataset. The bold text spans in two cases are entities. NPP, no relation between two medical problems; PIP, medical problem indicates medical problem.}\smallskip
	\begin{tabularx}{\linewidth}{cXp{6cm}p{0.75cm}p{1cm}p{0.5cm}p{0.7cm}}
		\hline & \bf Cases & \bf The Corresponding Triples & \bf BioBERT & \bf MedERNIE & \bf BERT-MK & \bf Ground Truth   \\ \hline
		1 & ... \textbf{coronary artery disease}, status post \textbf{mi} x0, cabg ... & (Coronary artery disease, associated with , MI) & NPP & PIP & PIP & PIP  \\ \hline
		2 & 0. \textbf{cad}: presented with \textbf{anginal symptoms} and ekg changes (stemi), with cardiac catheterization revealing lesions in lad, lcx, and plb. & (CAD, member of, Other ischemic heart disease (SMQ)); (Other ischemic heart disease (SMQ), has member, Angina symptom) & NPP & NPP & PIP & PIP   \\
		\hline 
	\end{tabularx}
\end{table*}

\paragraph{Effect of the Corpus Size in Pre-training}

Figure~\ref{fig:traindata} shows the model performance comparison with different proportion of the pre-training corpus.
From this figure, we observe that BERT-MK outperforms BioBERT by using only 10\%-20\% of the corpus, which indicates that medical knowledge has the capability to enhance pre-trained language models and save computational costs \cite{schwartz2019green}.

\section{Related Work}

Pre-trained language models represented by ELMO \cite{peters2018deep}, GPT \cite{radford2018improving} and BERT \cite{devlin2019bert} have attracted great attention, and a large number of variant models have been proposed.
Among these studies, some researchers devote their efforts  to introducing knowledge into language models, such as ERNIE-Baidu \cite{sun2019ernie} and ERNIE-Tsinghua \cite{zhang-etal-2019-ernie}.
ERNIE-Baidu introduces new masking units such as phrases and entities to learn knowledge information in these masking units.
As a reward, syntactic and semantic information from phrases and entities is implicitly integrated into the language model.
Furthermore, a different knowledge information is explored in ERNIE-Tsinghua, which incorporates knowledge graph into BERT to learn lexical, syntactic and knowledge information simultaneously.
However, information interaction between nodes in the knowledge graph is limited by the KRL method used in ERNIE-Tsinghua, which is crucial  to the knowledge quality.

Recently, several KRL methods based on neural networks have been proposed, which can be roughly divided into convolutional neural network (CNN) based \cite{dettmers2018convolutional,nguyen2018novel} and graph neural network (GNN) based models \cite{schlichtkrull2018modeling,nathani-etal-2019-learning}.
The CNN based models enhance the information interaction within a triple, but they treat a triple as a training unit and do not exploit the information of the relevant triples.
To overcome the above shortcoming, relational Graph Convolutional Networks (R-GCNs) \cite{schlichtkrull2018modeling} is proposed to learn entity embeddings from their incoming neighbors, which greatly enhances the information interaction between related triples.
\citeauthor{nathani-etal-2019-learning} \shortcite{nathani-etal-2019-learning} further extends the information flow from 1-hop in-entities to n-hop during the learning process of entity representations, and achieves the SOTA performance on multiple relation prediction datasets, especially for the ones containing higher in-degree nodes.

We believe that the information contained in knowledge graphs is far from being sufficiently exploited.
In this study, we develop a KRL method that can transform any subgraph into a training sample, which has the ability to model any information in the knowledge graph theoretically.
In addition, this knowledge representation is integrated into the language model to obtain an enhanced version of the medical pre-trained language model.

\section{Conclusion and Future Work}

We propose a novel approach to learn more comprehensive knowledge representation, focusing on modeling subgraphs in the knowledge graph by a Transformer-based method.
Additionally, the learned medical knowledge is integrated into the pre-trained language model, which outperforms BERT-Base and another two domain-specific pre-trained language models on several medical NLP tasks.
Our work validates the intuition that medical knowledge is beneficial to some medical NLP tasks and provides a preliminary exploration for the application of medical knowledge.

In the follow-up work, traditional downstream tasks for knowledge representation learning methods, such as relation prediction, will be used to further validate the effectiveness of KG-Transformer.
Moreover, we will explore a more elegant way to combine medical knowledge with language models.


%

\bibliographystyle{aaai}
\bibliography{aaai20}

\end{document}